\documentclass[letterpaper]{article} 
\usepackage[]{aaai23}  
\usepackage{times}  
\usepackage{helvet}  
\usepackage{courier}  
\usepackage[hyphens]{url}  
\usepackage{graphicx} 
\usepackage{natbib}  
\usepackage{caption} 
\usepackage{algorithm}
\usepackage{algorithmic}

\usepackage{newfloat}
\usepackage{listings}
\DeclareCaptionStyle{ruled}{labelfont=normalfont,labelsep=colon,strut=off} 
\floatstyle{ruled}
\newfloat{listing}{tb}{lst}{}
\floatname{listing}{Listing}
\usepackage{booktabs}

\newcommand{\eat}[1]{}

\let\oldhat\hat
\renewcommand{\vec}[1]{\mathbf{#1}}
\renewcommand{\hat}[1]{\oldhat{\mathbf{#1}}}
\renewcommand{\matrix}[1]{\mathbf{#1}}

\newcommand{\eg}{\emph{e.g.,}~}

\newcommand{\ie}{\emph{i.e.,}~}
\newcommand{\etc}{\emph{etc.}~}

\newcommand \footnoteONLYtext[1]
{
	\let \mybackup \thefootnote
	\let \thefootnote \relax
	\footnotetext{#1}
	\let \thefootnote \mybackup
	\let \mybackup \imareallyundefinedcommand
}
\title{STAGE: Span Tagging and Greedy Inference Scheme for Aspect Sentiment Triplet Extraction}
\author {
    Shuo Liang\textsuperscript{\rm 1,2},
    Wei Wei\textsuperscript{\dag,\rm 1,2},
    Xian-Ling Mao\textsuperscript{\rm 3},
    Yuanyuan Fu\textsuperscript{\rm 2,4},
    Rui Fang\textsuperscript{\rm 2,4},
    Dangyang Chen\textsuperscript{\rm 2,4}
}
\affiliations {
    \textsuperscript{\rm 1} Cognitive Computing and Intelligent Information Processing (CCIIP) Laboratory, \\School of Computer Science and Technology, Huazhong University of Science and Technology\\
    \textsuperscript{\rm 2} Joint Laboratory of HUST and Pingan Property \& Casualty Research (HPL)\\
    \textsuperscript{\rm 3} Department of Computer Science and Technology, Beijing Institute of Technology\\
    \textsuperscript{\rm 4} Ping An Property \& Casualty Insurance company of China, Ltd\\
    shuoliang@hust.edu.cn, weiw@hust.edu.cn, maoxl@bit.edu.cn,\\ fuyuanyuan83@gmail.com, fangrui051@pingan.com.cn, chendangyang273@pingan.com.cn
}

\begin{document}

\maketitle

\begin{abstract}
\footnoteONLYtext{\dag ~ Corresponding Author}
Aspect Sentiment Triplet Extraction~(ASTE) has become an emerging task in sentiment analysis research, aiming to extract triplets of the aspect term, its corresponding opinion term, and its associated sentiment polarity from a given sentence.
Recently, many neural networks based models with different tagging schemes have been proposed, but almost all of them have their limitations: heavily relying on 1) prior assumption that each word is only associated with a single role (\eg aspect term, \textit{or} opinion term, \etc) and 2) word-level interactions and treating each opinion/aspect as a set of independent words.
Hence, they perform poorly on the complex ASTE task, such as a word associated with multiple roles or an aspect/opinion term with multiple words.
Hence, we propose a novel approach, \underline{\textbf{S}}pan \underline{\textbf{TA}}gging and  \underline{\textbf{G}}reedy inf\underline{\textbf{E}}rence (STAGE), to extract sentiment triplets in span-level, where each span may consist of multiple words and play different roles simultaneously. 
To this end, this paper formulates the ASTE task as a multi-class \textbf{span} classification problem. 
Specifically, STAGE generates more accurate aspect sentiment triplet extractions via exploring span-level information and constraints, which consists of two components, namely, \textit{span tagging} scheme and \textit{greedy inference} strategy.
The former tag all possible candidate spans based on a newly-defined tagging set.
The latter retrieves the aspect/opinion term with the maximum length from the candidate sentiment snippet to output sentiment triplets.
Furthermore, we propose a simple but effective model based on the STAGE, which outperforms the state-of-the-arts by a large margin on four widely-used datasets.
Moreover, our STAGE can be easily generalized to other pair/triplet extraction tasks, which also demonstrates the superiority of the proposed scheme STAGE.

\end{abstract}

\section{Introduction\label{sec:intro}}
\begin{figure}[!t]
	\centering
	\includegraphics[width=0.45\textwidth]{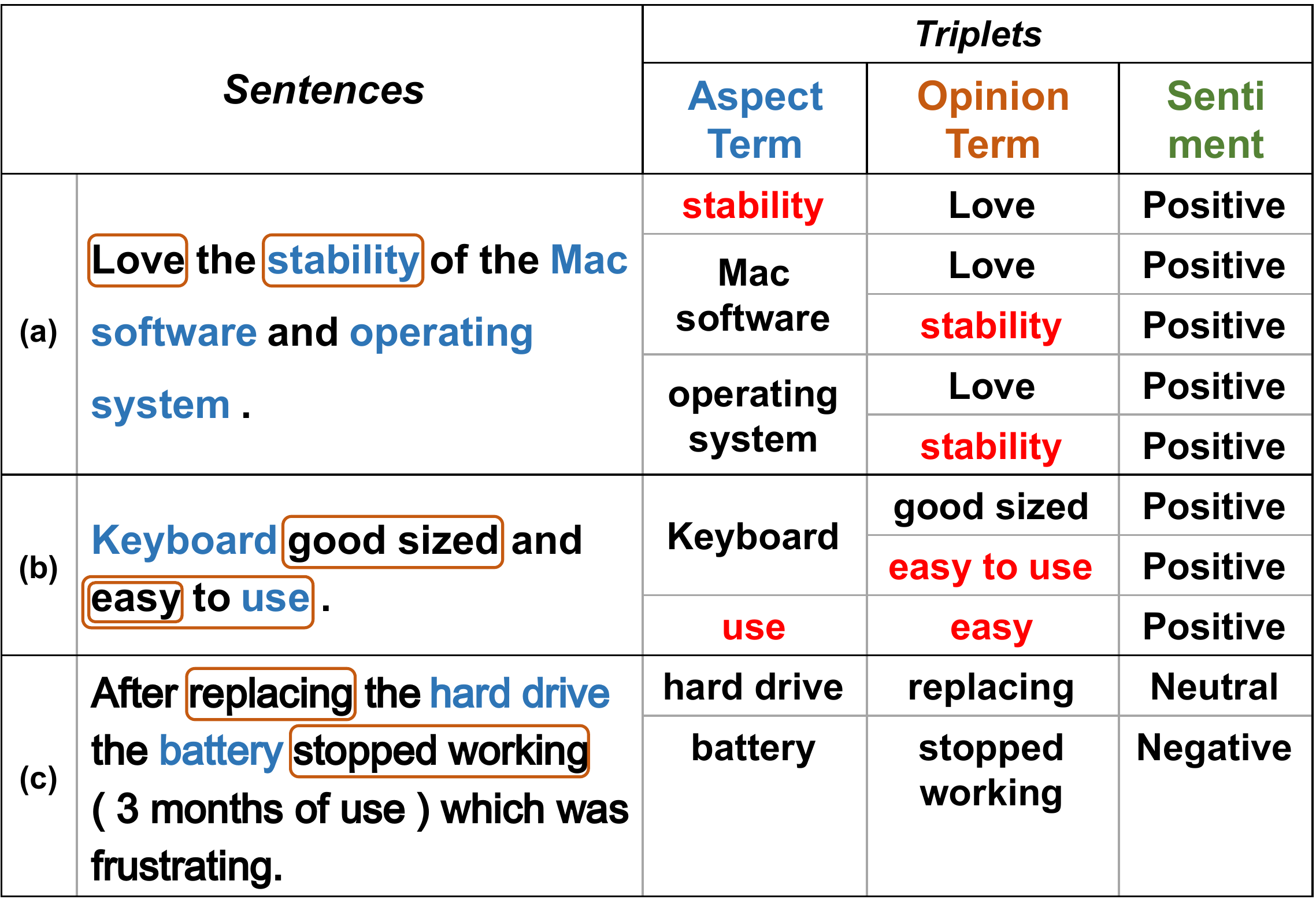}
	\caption{Examples of ASTE task from the real-world dataset. In each sentence, aspect terms are in blue and opinion terms are in rounded rectangles.} 
	\label{Fig:example_ASTE}
\end{figure}

Aspect Sentiment Triplet Extraction (ASTE) aims to identify all sentiment triplets from an input sentence, namely, the aspect term ($a$) with its corresponding opinion term ($o$) and sentiment polarity ($s$) (as shown in Figure~\ref{Fig:example_ASTE}).
%
It can be applied in many downstream tasks, such as dialogue generation~\cite{10.1145/3511808.3557359, 10.1145/3456414, 10.1145/3357384.3357937} and recommendation system~\cite{10.1145/3511808.3557404, 10.1145/3477495.3532025, zhao2022multi}.
Indeed, ASTE can be decomposed into several subtasks, such as 
Aspect Term Extraction (ATE)~\cite{li-lam-2017-deep, xu-etal-2018-double, dai-song-2019-neural}, 
Opinion Term Extraction (OTE)~\cite{fan-etal-2019-target, dai-song-2019-neural},
Aspect-Based Sentiment Classification (ABSC)~\cite{zhang-etal-2019-aspect, tang-etal-2020-dependency,liang-etal-2022-bisyn}, and
Aspect-Opinion Pair Extraction (AOPE) \cite{wang2017coupled, zhao-etal-2020-spanmlt}.
And a straightforward method is to independently extract elements of sentiment triplet through those subtasks in a pipeline fashion~\cite{peng2020knowing, chen2021bidirectional}.
However, the primary obstacle for those pipeline methods is the well-known error propagation problem.

Later, many efforts resort to extracting sentiment triplets in an end-to-end framework~\cite{xu-etal-2020-position, wu-etal-2020-grid, zhang-etal-2020-multi-task,yu2021aspect, xu-etal-2021-learning, mukherjee-etal-2021-paste, zhang-etal-2021-towards-generative, chen-etal-2022-enhanced,liu-etal-2022-robustly}.
Some works~\cite{mukherjee-etal-2021-paste, zhang-etal-2021-towards-generative} formulate it as a generative problem and achieve good performance under the benefit of existing generative pre-trained models, such as BART~\cite{lewis-etal-2020-bart} and T5~\cite{JMLR:v21:20-074}.
In addition, most of end-to-end approaches focus on designing a new tagging scheme, converting ASTE into a classification problem towards each \textit{word}~\cite{xu-etal-2020-position} or each \textit{word pair}~\cite{wu-etal-2020-grid, chen-etal-2022-enhanced}.
Despite their success, those \textit{word-level} tagging schemes faces several problems: heavily relying on 
(1) \textbf{Prior assumption that restricts the diversity of word roles.} Those tagging schemes only work when each word is only associated with a single role~(the aspect, \textit{or} the opinion, \etc) \textit{or} when each word pair corresponds to one specific relation. As a result, they cannot perform well when it comes to a word playing multiple roles or a word pair having multiple relations.
(2) \textbf{Word-level interactions, thus falling short of utilizing span-level information.} 
Those methods only focus on the interactions between words and treat the multi-word term as a set of independent words, thus separating the semantic information of the whole span when facing multi-word aspect/opinion terms. It also causes difficulty in guaranteeing the sentiment consistency when extracting triplets.
Without loss of generality, we illustrate those problems with examples in Figure~\ref{Fig:example_ASTE}:
(1) \textbf{\textit{Multiple roles of a word.}} ``Stability'' from sentence (a) has two roles simultaneously, that is, being (part of) the aspect term and the opinion term, which confuses the \textit{word-level} tagging schemes when figuring out the proper tag;
(2) \textbf{\textit{Multiple relations of a word-pair.}} In sentence (b), not only ``easy'' and ``use'' are words of the same opinion term (\ie easy to use), but they form a valid aspect-opinion pair as well, resulting in multiple relations of the word pair ``easy-use'', which also challenges the existing tagging scheme;
(3) \textbf{\textit{Multi-word term.}} It is common that an aspect/opinion term contains multiple words, and a significant semantic difference may exist between the whole term and the words in it, such as  ``hard drive'' from sentence (c). 
Treating those multi-word terms as a set of independent words, the \textit{word-level} tagging schemes fail to utilize their complete semantics and consequently deteriorate the performance of the extraction task. 
Moreover, in \textit{word-level} tagging methods, words that make up a term are classified independently and thus may contain different sentiment polarities.
Therefore, extra effort is required to deal with this sentiment inconsistency problem when outputting triplets.

The aforementioned problems motivate us to explore a \textit{span-level} tagging method and formulate it as a multi-class classification problem towards each \textit{span}.
Generally, it is observed that a span in the given sentences may have three roles: aspect term, opinion term, and a snippet that includes an aspect term and the corresponding opinion term, forming its boundaries (\ie sentiment snippet), such as ``\underline{love} the \underline{stability}'' from Figure~\ref{Fig:example_ASTE} (a). Note that a span can play different roles simultaneously.

Based on the above viewpoint, we pre-define three role dimensions~(\ie aspect term, opinion term, and sentiment snippet) and propose a novel \textit{span-level} tagging-based approach, \underline{\textbf{S}}pan \underline{\textbf{TA}}gging and \underline{\textbf{G}}reedy inf\underline{\textbf{E}}erence~(STAGE), to solve ASTE in an end-to-end fashion. By exploring span-level information and constraints, STAGE can efficiently generate more accurate sentiment triplets.
Specifically, it consists of two components: 
(1) \textit{Span tagging} scheme: which tags all possible spans based on the defined role dimensions, guaranteeing the diversity of span roles. Thus, it can depict a complete picture of each span, like how many and what roles a span plays.
(2) \textit{Greedy inference} strategy: which decodes all triplets simultaneously in a more accurate and efficient way by (i) considering the mutual span constraints between the sentiment snippets and aspect/opinion terms and (ii) only retrieving the aspect/opinion term with the maximum length from a sentiment snippet, compared to previous decoding algorithms that look into every possible pair combination of candidate aspects and opinions to see whether they are valid.
Therefore, naturally performing at the span-level, STAGE can overcome the problems of existing \textit{word-level} tagging schemes.
Based on STAGE, we propose a simple model to demonstrate its effectiveness.

Moreover, with a minor tagging set change, STAGE is very easily generalized to other pair/triplet extraction tasks, such as entity and relation extraction~\cite{zhong-chen-2021-frustratingly,yan-etal-2021-partition}, a fundamental task aiming to extract (head entity, tail entity, relation) triplets from the given sentence\footnote{More details are shown in Section\ref{section:analysis} Analysis.}.

In summary, the key contributions are as follows:

(1) To the best of our knowledge, we make the first effort to explore the \textit{span-level} tagging method and formulate the ASTE task as a multi-class \textbf{span} classification problem.

(2) We propose a novel end-to-end tagging-based approach, \underline{\textbf{S}}pan \underline{\textbf{TA}}gging and \underline{\textbf{G}}reedy inf\underline{\textbf{E}}erence~(STAGE). Specifically, it consists of the \textit{span tagging} scheme that considers the diversity of span roles, overcoming the limitations of existing tagging schemes, and the \textit{greedy inference} strategy that considers the span-level constraints, generating more accurate triplets efficiently.

(3) We propose a simple but effective model based on STAGE and extensive experiments on four widely-used English datasets verify its effectiveness. Furthermore, it can be easily generalized to other pair/triplet extraction tasks, which also demonstrates the superiority of STAGE.
\section{Related Work}
We present some commonly used tagging schemes in the ASTE task, which can be categorized into two main kinds.

\subsection{Sequence Tagging}
The base tagging scheme is BIESO tagging\footnote{BIESO means ``begin, inside, end, single, other'', respectively.}. 
Most pipeline models firstly adopt it to extract aspects and opinions separately and then determines whether any two of them can form a valid aspect-opinion pair with the corresponding sentiment. 
Generally, they view ASTE as two sequence labeling problems (\ie classification problems towards each \textit{word}) plus one classification problem towards each candidate aspect-opinion pair. 
\citet{peng2020knowing} propose a two-stage pipeline approach and utilize a unified tagging scheme to tag the boundary of aspect term with its sentiment and a common BIESO tagging scheme to extract opinion term.

However, the pipeline approaches ignore the interactions between three triplet elements and commonly suffer from error propagation.
Besides, those tagging schemes encode no positional information and thus fail to model rich interactions among aspect and opinion terms. 
So \citet{xu-etal-2020-position} propose JET, an extended BIESO tagging scheme with position index information. However, it still fails to simultaneously tackle the situations that one aspect/opinion term relates to multiple opinion/aspect terms.

\subsection{Table Tagging}
The basic idea is to tag a $|n|\times|n|$ table $T$, where $n$ is the length of the input text and $T[i][j]$ corresponds to the relation between $i$-th word and $j$-th word.
\citet{wu-etal-2020-grid} firstly propose a grid tagging scheme, utilizing a tag set \{A, O, POS, NEU, NEG, N\} to donate five possible relations of a word-pair. 
In order to utilize more relations between words, \citet{chen-etal-2022-enhanced} define ten relations and consider diverse linguistic features.
The above approaches both perform in an end-to-end way, avoiding error prorogation and can simultaneously handle the case of one aspect term corresponding to multiple opinion terms and vice versa. 
However, as we mentioned before, they still face several limitations.
%
%
\section{Methodology}
In this section, we first introduce the ASTE task and then explain our proposed span tagging and greedy inference (STAGE) scheme in detail. Finally, we present our implemented model. 

\subsection{Problem Statement}
 Given the input sentence $X = \left\{w_1, w_2, ..., w_n\right\}$, where $w_i$ denotes a word and $n$ is the sentence length. The goal of ASTE task is to extract a set of sentiment triplets $\mathcal{T}=\left\{(a,o,s)_m\right\}_{m=1}^{|\mathcal{T}|}$ from the sentence $X$, where $a$, $o$ and $s$ represent aspect term, opinion term and sentiment polarity separately and $s$ $\in$ \{POS, NEG, NEU\}.
 We use $SP = \left\{SP_{1,1},SP_{1,2},...,SP_{i,j},...,SP_{n,n}\right\}$ to illustrate all possible enumerated spans in $X$, where $i$ and $j$ represent the start and end positions in $X$, respectively.
 So the formal definition for the sentiment snippet is as follows,
 $SP_{i,j}$ is an sentiment snippet menas ($SP_{i,k}$, $SP_{l,j}$) is a valid  (aspect, opinion) or (opinion, aspect) pair, where $i \le k,l \le j$. 

\begin{table}[t]\small
		\centering
		\begin{tabular}{cc}
				\toprule[1.5pt]
				Sub Tags & Meaning \\
				\hline
				A & an aspect term \\
				\hline
				O & an opinion term \\
				\hline
				NEG & a sentiment snippet containing \\
					& \textit{negative} aspect-opinion pair.\\
				\hline
				NEU & a sentiment snippet containing \\
					& \textit{neutral} aspect-opinion pair.\\
				\hline
				POS & a sentiment snippet containing \\
				 	& \textit{positive} aspect-opinion pair. \\
				\hline
				N   & nothing in the specific dimension \\
				\bottomrule[1.5pt]
			\end{tabular}
		\caption{\label{Table:tag_scheme} Descriptions of sub tags.}
	\end{table}

\begin{figure}[tb]
	\centering
	\includegraphics[width=0.47\textwidth]{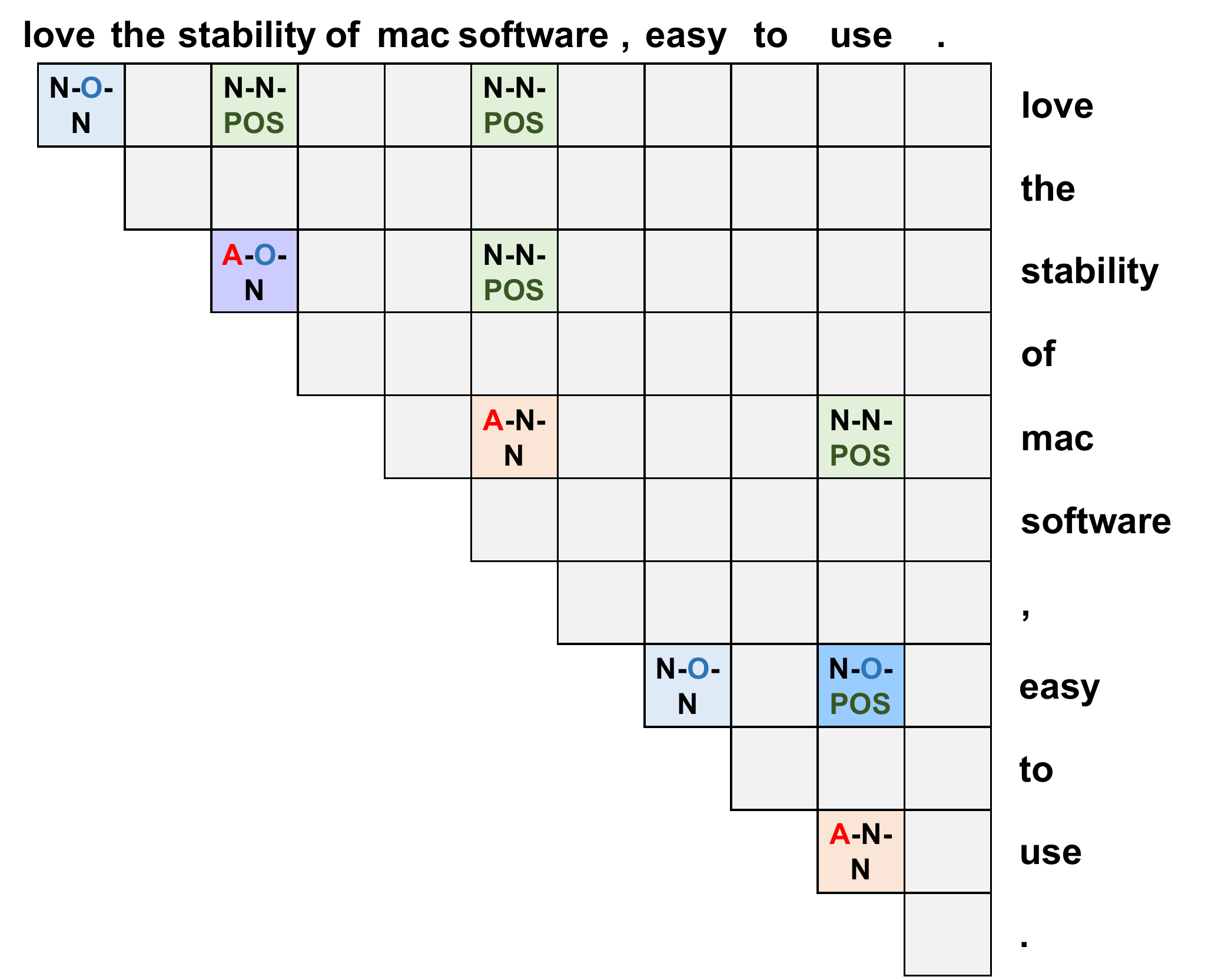}
	\caption{Example of 3D-version span tagging. Cells with gray background represent ``N-N-N''.} 
	\label{Fig:tagging_example}
\end{figure}

 \subsection{STAGE: \underline{S}pan \underline{TA}gging and \underline{G}reedy inf\underline{E}rence}
\subsubsection{Span Tagging}
Considering the diversity of span roles,
our span tagging scheme depict each span~(\ie $SP_{i,j}$) in three role dimensions: whether $SP_{i,j}$ is a valid {aspect term}, a valid {opinion term} and a valid {sentiment snippet}. We use $role^a_{i,j}$, $role^o_{i,j}$, $role^s_{i,j}$ to represent them separately, where $role^a_{i,j} \in$ \{N,A\}, $role^o_{i,j} \in$ \{N,O\}, $role^s_{i,j} \in$ \{N, NEG, NEU, POS\}. 
The meanings of those sub tags are shown in Table~\ref{Table:tag_scheme}. 

By considering those three dimensions independently, we propose a \textbf{3D-version} span tagging method with the tag set \{N, A\}-\{N, O\}-\{N, NEG, NEU, POS\}.
The tagging results are like an upper triangular table $T$, where $T[i][j]$ represents the tag for span $SP_{i,j}$, as illustrated by Figure~\ref{Fig:tagging_example}.
Note that the tag for span ``easy to use'' is ``{N-O-POS}'' because the span is not only an opinion term but also contains ``(use, easy)'', a valid aspect-opinion pair with positive sentiment. 


%
Apart from considering the three dimensions independently, we also investigate two variants:

1) \textbf{2D-version}: With tag set \{N, O, A\}-\{N, NEG, NEU, POS\}, it can be viewed as projecting the aspect and opinion role dimensions into one. The tag ``N'' in the first dimension denotes the span is neither an aspect nor an opinion. This variant cannot handle the situations that a span being aspect and opinion simultaneously.

2) \textbf{1D-version}: With tag set \{N, NEG, NEU, POS, O, A\}, it can be viewed as projecting the three role dimensions into one. The tag ``N'' means the span plays no roles. This variant fails to deal with the cases that a span has multiple roles.

\begin{algorithm}[!t]\normalsize
	\caption{ Greedy Inference.}
	\label{Algo:greedy_inference}
	\renewcommand{\algorithmicrequire}{ \textbf{Input:}}
	\renewcommand{\algorithmicensure}{ \textbf{Output:}} 

	\begin{algorithmic}[1] 
		\REQUIRE ~~\\ 
		The tagging results $P$ of the sentence $X$ with length $n$.
		 $P_{i,j}$ represents the tag label of the span $SP_{i,j}$  and can be decomposed into three dimensions $role^a_{i,j}$, $role^o_{i,j}$ and $role^s_{i,j}$.
		
		\ENSURE ~~\\ 
		Triplets $\mathcal{T}$ of the given sentence;
		\STATE Initialize $\mathcal{A}=\{\}$, $\mathcal{O}=\{\}$, $\mathcal{D} = \{\}$, $\mathcal{T} = \{\}$;
		\STATE $\mathcal{A} = \{(i,j) | role^a_{i,j} = $ A$, 0 \le i \le j \le n\}$
		\STATE $\mathcal{O} = \{(i,j) | role^o_{i,j} = $ O$, 0 \le i \le j \le n\}$
		\STATE $\mathcal{D} = \{(i,j, role^s_{i,j}) | role^s_{i,j} \neq $ N$, 0 \le i \le j \le n\}$

		\FOR{each $(i,j,s)$ in $\mathcal{D}$}
		\STATE $//$ CASE-1: aspect term before opinion term;
		\STATE $ CA = \{ k \ | i \le k \le j, \ (i,k) \in \mathcal{A}\}$
		\STATE $ CO = \{ l\ \ | i \le l\ \le j, \ (l,j) \in \mathcal{O}\}$
		\IF {$CA \neq \emptyset$ and $CO \neq \emptyset$}
		\STATE remove $j$ (if exists) from $CA$ when $|CA| > 1$ 
		\STATE remove $i$ (if exists) from $CO$ when $|CO| > 1$
		\STATE $k = max(CA)$, $l=min(CO)$
		\STATE 
		$\mathcal{T} \leftarrow \mathcal{T} \cup (SP_{i,k},~SP_{l,j},~s)$
		\ENDIF
		
		\STATE $//$ CASE-2: aspect term after opinion term;
		\STATE $ CO = \{ k\ | i \le k \le j, \ (i,k) \in \mathcal{O} \}$
		\STATE $ CA = \{ l\ \ | i \le l\ \le j,\  (l,j) \in \mathcal{A} \}$

		\IF {$CO \neq \emptyset$ and $CA \neq \emptyset$}
		\STATE remove $j$ (if exists) from $CO$ when $|CO| > 1$ 
		\STATE remove $i$ (if exists) from $CA$ when $|CA| > 1$
		\STATE $k = max(CO)$, $l=min(CA)$
		\STATE $\mathcal{T} \leftarrow \mathcal{T} \cup (SP_{l,j},~ SP_{i,k},~s)$
		\ENDIF
		
		\ENDFOR
		\RETURN $\mathcal{T}$; 
	\end{algorithmic}
\end{algorithm}

\subsubsection{Greedy Inference}
Given the tagging results in our span tagging scheme, in order to generate more accurate triplets efficiently, we propose a greedy inference strategy, where the ``\textbf{greedy}'' means retrieving the aspect/opinion term with the maximum length separately in a valid sentiment snippet. 

The details are shown in Algorithm~\ref{Algo:greedy_inference}. Firstly, we obtain all possible aspects, opinions and sentiment snippets from the tagging results~(line 1 to line 4). Then we look into every sentiment snippet~(line 5 to line 24) and consider two cases: 

(1) The aspect term appears before the opinion term~(line 7 to line 14), which means the former shares the start index and the latter shares the end index with the current sentiment snippet~(line 7 to line 8). 
If both candidate aspects and opinions exist, we then separately retrieve the one with the maximum length, forming a valid triplet~(line 9 to line 14).
Considering that the sentiment snippet itself can be the aspect/opinion, we only view it as an aspect/opinion when there is no other counterpart candidates~(line 10/11);

(2) The aspect term appears after the opinion term (line 16 to line 23), which is very similar to the previous case, except the candidate aspects share the end index and the candidate opinions share the start index with current sentiment snippet.

Finally, we can obtain all valid sentiment triplets after examining all sentiment snippets.
Compared to previous decoding algorithms, our strategy has advantages in:

(1) \textbf{Accuracy}: 
The mutual span-level constraints guarantee the accuracy of the extracted sentiment triplets.
For the sentiment snippet, which also indicates the pairing of the aspect and opinion, not only itself should have correct sentiment polarity, but also it should contain valid aspects and opinions that simultaneously sharing one of its boundaries. 
For the aspect/opinion term, not only themselves should be classified correctly, but also the boundaries of them should form a valid sentiment snippet. 

(2) \textbf{Efficiency}: 
Contrary to the common  practice that looks into every possible pair combination of candidate aspects and opinions to see whether they are valid, which contributes to heavy computational overhead, our method extracts the triplets based on the predicted sentiment snippets, the number of which is much less. More specifically, the time complexity is $O(n^2)$ with greedy inference and $O(n^4)$ with common practice, where $n$ is length of input text.

\begin{figure}[tb]
	\centering
	\includegraphics[width=0.36\textwidth,height=0.42\textwidth]{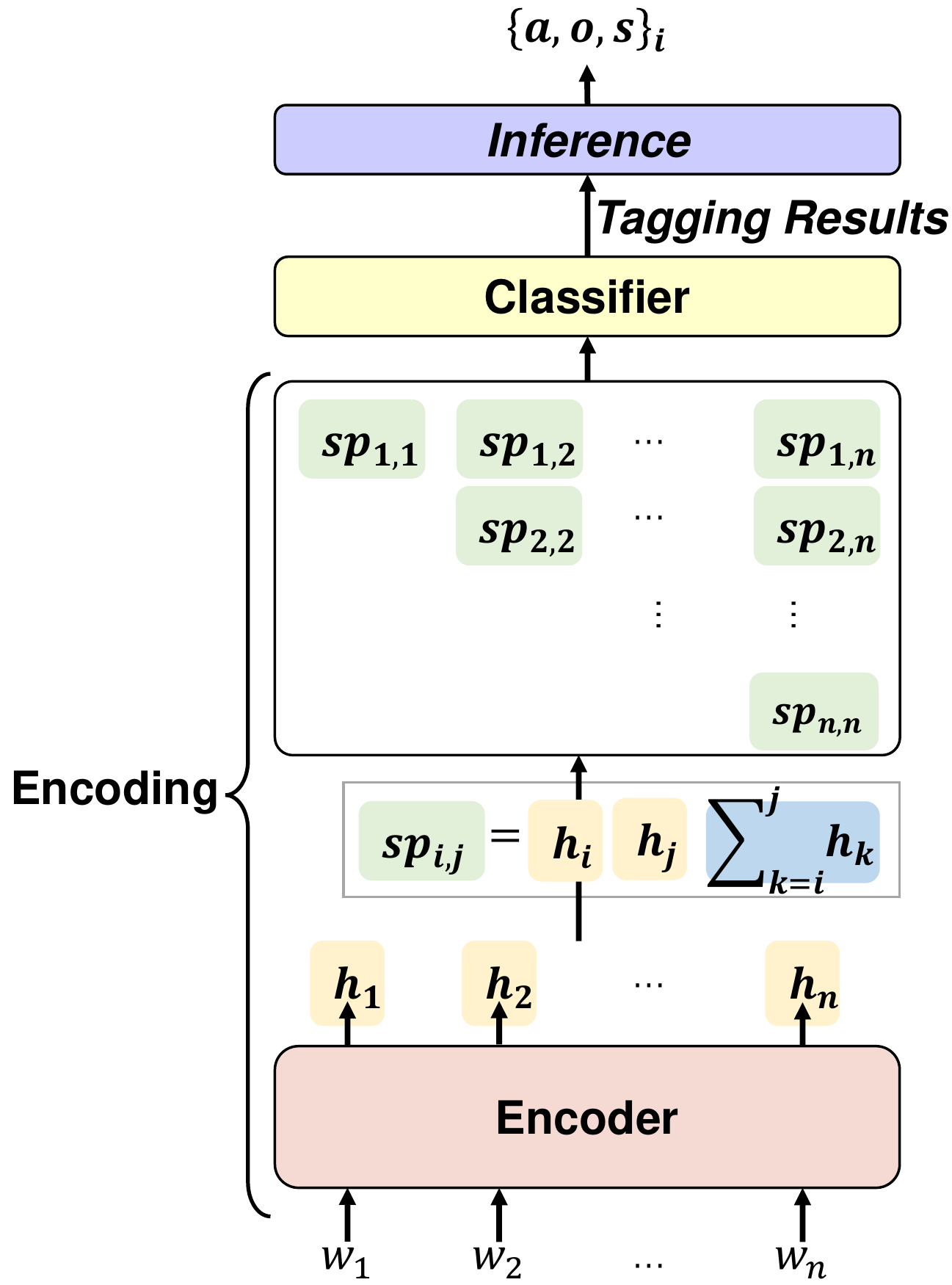}
	\caption{Model Architecture. } 
	\label{Fig:model}
\end{figure}

\begin{table*}[htb]\footnotesize
	\centering
	\begin{tabular}{cc|cc|ccc|ccc|ccc|c}
		\toprule
		\multicolumn{2}{c|}{\textbf{Dataset}} & \textbf{\#S} & \textbf{\#T} & \textbf{\#Pos.} & \textbf{\#Neu.} & \textbf{\#Neg.} & \multicolumn{7}{c}{\textbf{Role distributions of spans}~(aspect-opinion-sentiment snippet)} \\
		\cline{8-14} 
		& & & & & & & A-N-N & N-O-N & N-N-S & A-O-N & A-N-S & N-O-S & A-O-S \\ 
		\hline
		~ 	  & Train & 906  & 1460 & 817  & 126 & 517 & 1279 & 1263 & 1454 & 1 & 1 & 4  & 0 \\
		14lap & Dev   & 219  & 345  & 169  & 36  & 140 & 296  & 304  & 344  & 0 & 0 & 0  & 0 \\
		~ 	  & Test  & 328  & 541  & 364  & 63  & 114 & 463  & 473  & 540  & 0 & 0 & 1  & 0 \\
		\hline
		~     & Train & 1266 & 2337 & 1691 & 166 & 480 & 2043 & 2072 & 2313 & 0 & 7 & 13 & 1 \\
		14res & Dev   & 310  & 577  & 404  & 54  & 119 & 497  & 498  & 569  & 0 & 3 & 5  & 0 \\
		~     & Test  & 492  & 994  & 773  & 66  & 155 & 840  & 850  & 984  & 1 & 7 & 3  & 0 \\
		\hline
		~     & Train & 605  & 1013 & 783  & 25  & 205 & 861  & 941  & 1012 & 0 & 1 & 0  & 0 \\
		15res & Dev   & 148  & 249  & 185  & 11  & 53  & 213  & 236  & 249  & 0 & 0 & 0  & 0 \\
		~     & Test  & 322  & 485  & 317  & 25  & 143 & 432  & 461  & 485  & 0 & 0 & 0  & 0 \\
		\hline
		~     & Train & 857  & 1394 & 1015 & 50  & 329 & 1197 & 1307 & 1393 & 0 & 1 & 0  & 0 \\
		16res & Dev   & 210  & 339  & 252  & 11  & 76  & 296  & 319  & 339  & 0 & 0 & 0  & 0 \\
		~     & Test  & 326  & 514  & 407  & 29  & 78  & 451  & 473  & 513  & 1 & 0 & 1  & 0 \\
		\bottomrule
	\end{tabular}
	\caption{\label{Table:dataset} Statistics of datasets. \#S, \#T mean the total number of sentences and triplets. \#Pos., \#Neu. and \#Neg. denote the number of positive, neutral, and negative sentiment triplets respectively. Note that we remove the duplicate ones. Role distributions of spans show the number of the spans playing different roles from three dimensions~(aspect, opinion and sentiment snippet), where N means nothing in that dimension, A, O and S denote aspect,opinion and sentiment snippet, respectively.}
\end{table*}

\subsection{STAGE Model}
Based on STAGE, we implement a simple BERT-based model. Shown in Figure~\ref{Fig:model}, the overall model architecture contains an encoder that outputs contextual word representations for generating span representations and a classifier that predicts the tag of each span. The inference module is used to generate triplets from the tagging results. 

\subsubsection{Encoding} The key point of this process is to obtain span representations. Firstly, we adopt BERT~\cite{devlin-etal-2019-bert} plus an additional fully connected layer as our encoder to generate contextual word representations $\vec{h}_{i}$ by,
\begin{equation}
	\matrix{\hat{h}} = BERT([CLS] + \{w_i\} +[SEP]) ,
\end{equation} 
\begin{equation}
	\vec{h}_{i}= \matrix{W_{fc}} (\dot \frac{1}{|BertT(w_i)|}\sum_{k \in BertT(w_i) }\vec{\hat{h}}_k) + \vec{b_{fc}},
\end{equation}
where $BertT(w_i)$ returns the index set of $w_i$'s sub-words in BERT sequence, and $|\cdot|$ returns its length. $\matrix{W_{fc}}$ and $\vec{b_{fc}}$ are the parameters of the fully connected layer, which is used to project the word representations from BERT to a lower dimension space. Then we form the span representations by,
\begin{equation}
	\vec{sp}_{i,j}= \vec{h}_i \oplus \vec{h}_j \oplus \sum_{k=i}^{j}\vec{h}_k,
\end{equation}
where $\oplus$ means vector concatenation. In this way, the model can utilize both the boundary information and the whole semantics of the span.

\subsubsection{Training}
The span representations are fed to a fully connected layer (\ie classifier) with a softmax activation function, generating the probabilities over role tags:
\begin{equation}
	\vec{p(sp_{i,j})} = softmax(\matrix{W_r}\vec{sp_{i,j}}+\vec{b_r}),
\end{equation}
where $\matrix{W_p}$, $\vec{b_p}$ are parameters of the classifier.

The training loss is defined as the cross-entropy loss between golden labels and predicted label distributions of all spans:
\begin{equation}
	L (\theta) =-\sum_{i=1}^{n} \sum_{j=i} ^{n} loss(\vec{p(sp_{i,j})},y_{i,j}),
\end{equation}
where $y_{i,j}$ represent the gold label for $SP_{i,j}$, $loss$ is the standard cross-entropy loss and $\theta$ represents model parameters.

\begin{table*}[!t] \footnotesize
	\centering
	\begin{tabular}{c|ccc|ccc|ccc|ccc}
		\toprule
		\textbf{Model} & \multicolumn{3}{c|}{\textbf{14lap}} & \multicolumn{3}{c|}{\textbf{14res}} & \multicolumn{3}{c|}{\textbf{15res}} & \multicolumn{3}{c}{\textbf{16res}} \\
		
		~ & $P$ & $R$ & $F_1$ & $P$ & $R$ & $F_1$ &$P$ & $R$ & $F_1$ & $P$ & $R$ & $F_1$  \\
		\midrule[1pt]
		CMLA+$^\sharp$ 	        & 30.09 & 36.92 & 33.16 
		& 39.18 & 47.16 & 42.79 
		& 34.56 & 39.84 & 37.01 
		& 41.34 & 42.10 & 41.72 \\
		
		RINANTE+$^\sharp$ 	    & 21.71 & 18.66 & 20.07 
		& 31.42 & 39.38 & 34.95 
		& 29.88 & 30.06 & 29.97 
		& 25.68 & 22.30 & 23.87 \\
		
		Li-unified-R$^\sharp$   & 40.56 & 44.28 & 42.34 
		& 41.04 & 67.35 & 51.00 
		& 44.72 & 51.39 & 47.82 
		& 37.33 & 54.51 & 44.31 \\
		
		Peng-two-stage$^\sharp$ & 37.38 & 50.38 & 42.87 
		& 43.24 & 63.66 & 51.46 
		& 48.07 & 57.51 & 52.32 
		& 46.96 & 64.24 & 54.21 \\
		
		OTE-MTL~$^\flat$		& 49.53 & 39.22 & 43.42 
		& 62.00 & 55.97 & 58.71
		& 56.37 & 40.94 & 47.13 
		& 62.88 & 52.10 & 56.96  \\
		
		JET-BERT~$^\sharp$		& 55.39 & 47.33 & 51.04 
		& 70.56 & 55.94 & 62.40 
		& 64.45 & 51.96 & 57.53 
		& 70.42 & 58.37 & 63.83  \\
		
		GTS-BERT~$^\dag$ 		& 58.62 & 52.35 & 55.29  
		& 68.14 & 68.77 & 68.45 
		& 62.37 & 59.71 & 60.98 
		& 66.16 & 68.81 & 67.44  \\
		
		BMRC~$^\flat$ 			& 70.55 & 48.98 & 57.82 
		& 75.61 & 61.77 & 67.99 
		& 68.51 & 53.40 & 60.02 
		& 71.20 & 61.08 & 65.75  \\
	

		ASTE-RL	& 64.80 & 54.99 & 59.50 
		& 70.60 & 68.65 & 69.71 
		& 65.45 & 60.29 & 62.72 
		& 67.21 & 69.69 & 68.41  \\
		
		Span-ASTE~$^\dag$ 		& 62.74 & 56.75 & 59.58 
		& 74.17 & 68.27 & 71.00 
		& 64.15 & 62.10 & 63.05 
		& 69.31 & 71.32 & 70.29  \\
		
		EMC-GCN~$^\dag$ &	59.61 & 56.30 & 57.90 
		& 70.37 & 72.84 & 71.58 
		& 60.45 & 62.72 & 61.55
		& 63.43 & 72.63 & 67.69  \\
%
		
		
		\midrule[1pt] 
		
		STAGE-1D 	& 71.48 & 53.97 & 61.49
		& 79.54 & 68.47 & 73.58
		& 72.05 & 58.23 & 64.37
		& 78.38 & 69.10 & \textbf{73.45} \\

		STAGE-2D  & 70.56 & 55.16 & \textbf{61.88}
		& 78.51 & 69.30 & \underline{73.61}  
		& 72.33 & 58.93 & \textbf{64.94} 
		& 77.67 & 68.44 & 72.75 \\

		STAGE-3D  & 71.98 & 53.86 & \underline{61.58}
		& 78.58 & 69.58 & \textbf{73.76}
		& 73.63 & 57.90 & \underline{64.79}
		& 76.67 & 70.12 & \underline{73.24} \\

		\bottomrule
		
	\end{tabular}
	\caption{\label{Table:Exp}
		Experimental results on the test set. The best $F_1$ results are in bold. The second best $F_1$ results are underlined. The ``$^\sharp$'' and  ``$^\flat$'' mean that the results are retrieved from \citet{xu-etal-2021-learning} and \citet{chen-etal-2022-enhanced}, respectively. The ``$^\dag$'' means we reproduce the results using the official released implementation code and configuration.
	}
\end{table*}

\section{Experiment}
\subsection{Datasets and Metrics}
To evaluate our proposed STAGE model, we conduct experiments on four commonly-used datasets released by ~\citet{xu-etal-2020-position}\footnote{ASTE-Data-V2 from \url{https://github.com/xuuuluuu/SemEval-Triplet-data}} that includes one dataset in laptop domain and three datasets in restaurant domain.
It is a refined version of the dataset released by~\citet{peng2020knowing} as the latter contains unannotated triplets. 
Those datasets are derived from the SemEval Challenge~\cite{Pontiki2014SemEval2014T4, Pontiki2015SemEval2015T1, Pontiki2016SemEval2016T5} and the annotations of the opinion terms are retrieved from~\citet{fan-etal-2019-target}.
Some detailed statistics are shown in Table~\ref{Table:dataset}. 
The numbers in the A-O-N, A-N-S, N-O-S and A-O-S columns indicate that it is common that a span plays different roles simultaneously in the sentence, implying that a word may have multiple roles and a word-pair may have multiple relations, which contradicts the assumption that previous tagging schemes~\cite{wu-etal-2020-grid,chen-etal-2022-enhanced} are based.
Following previous works, ``Precision($P$)'', ``Recall($R$)'' and ``$F_1$ scores ($F_1$)'' are the evaluation metrics. 

\subsection{Implementation Details}
The proposed STAGE\footnote{Code is available at https://github.com/CCIIPLab/STAGE} model contains a BERT-base-uncased model\footnote{https://github.com/huggingface/transformers} and a fully connected layer as encoder, with hidden state dimension of 768 and 200 respectively. The dropout rate is 0.5. The model is trained in 100 epochs with a batch size of 16.
AdamW optimizer~\cite{Loshchilov2017FixingWD} is adopted with a learning rate of $2 \times 10 ^{-5}$ for BERT fine-tuning and $ 10 ^{-3}$ for other trainable parameters.
For each dataset, we select the model with the best $F_1$ scores on the development set and report the average results of five runs with different random seeds.

\subsection{Baselines}
We compare our model with the following baselines:

$\bullet$~Pipeline:
\textbf{CMLA+}~\cite{wang2017coupled} considers the aspect-opinion interactions by an attention mechanism,
\textbf{RINANTE+}~\cite{dai-song-2019-neural} uses a BiLSTM-CRF and models the dependency relations, and \textbf{Li-unified-R}~\cite{li2019unified} adopts a unified tagging method to jointly extract the aspect term and its sentiment, all modified by~\citet{peng2020knowing} to perform ASTE task.
\textbf{Peng-two-stage}~\cite{peng2020knowing} models the interaction between aspects and opinions. 

$\bullet$~End-to-End: 
\textbf{OTE-MTL}~\cite{zhang-etal-2020-multi-task} adopts a multi-task learning framework with a biaffine scorer.
\textbf{JET-BERT}~\cite{xu-etal-2020-position} introduces position information into the sequence tagging scheme.
\textbf{GTS-BERT}~\cite{wu-etal-2020-grid} performs ASTE task by predicting the sentiment relations of word pairs.
\textbf{Span-ASTE}~\cite{xu-etal-2021-learning} learns span interactions between aspects and opinions.
\textbf{BMRC}~\cite{chen2021bidirectional} views ASTE as an multi-turn machine reading comprehension problem.
\textbf{ASTE-RL}~\cite{yu2021aspect} adopts a hierarchical reinforcement learning framework by considering the aspect and opinion terms as arguments of the expressed sentiment.
\textbf{EMC-GCN}~\cite{chen-etal-2022-enhanced} proposes to fully utilize the relations between words and diverse linguistic features.

\subsection{Overall Performance}
The overall performances are shown in Table~\ref{Table:Exp}. The observations are that:
(1) Under the $F_1$ scores, our STAGE models (including two variants) all outperform state-of-the-art baselines by large margins in four datasets.
And compared to previous best results, STAGE-3D significantly exceeds their $F_1$ scores by an average of 2.22\%. And two variants (STAGE-2D and STAGE-1D) also exceed by an average of 2.17\% and 2.10\%, respectively.
Considering our model architecture is quite simple, those significant improvements can well verify the effectiveness of our STAGE scheme.
(2) Compared with STAGE-3D, the variant STAGE-2D achieves competitive results, even though it cannot deal with the cases that a span is the aspect and opinion simultaneously. 
It may result from the very limited training samples in those cases (as shown in Table~\ref{Table:dataset}), which causes STAGE-3D cannot learn such knowledge well. 
(3)  STAGE-1D exceeds GTS-BERT~\cite{wu-etal-2020-grid} by an average 5.18\% under the $F_1$ score. 
Considering that the two models have the same tag set ~(\ie \{A, O, POS, NEU, NEG, N\}) but with different tagging schemes, it verifies that \textit{span-level} tagging is superior to \textit{word-level} tagging. 
Moreover, all our models outperform two state-of-the-art tagging-based baselines GTS-BERT and EMC-GCN significantly and consistently.
(4) Although exceeding most baselines, our recall results are slightly lower than the best. But our precision results are better. One reason is that considering the mutual span-level constraints makes the process of retrieving and pairing much stricter during the inference.

\begin{table}[!t]\footnotesize
	\centering
	\begin{tabular}{c|c|cccc}
		\toprule
		\textbf{Task} & 
		\textbf{Model} & \textbf{14lap} & \textbf{14res} & \textbf{15res} & \textbf{16res} \\
		\midrule[1pt]
		& GTS-BERT  & 80.44  & 82.76  & 78.27 & 81.13 \\
		& EMC-GCN   & 81.67  & 84.68  & 77.62 & 80.57 \\
		\cline{2-6}
		ATE 	& STAGE-1D  & \textbf{82.96} & \textbf{85.24} & 80.91 & 83.60 \\
		& STAGE-2D  & \underline{82.64}	& \underline{85.23} & \textbf{81.39} & \textbf{83.95} \\
		& STAGE-3D 	& 82.62 & 84.99 & \underline{81.21} & \underline{83.86} \\
		\midrule[1.5pt]
		& GTS-BERT	& 77.82  & 84.85 & 77.53 & 84.36\\
		& EMC-GCN   & 78.85  & 85.62 & 78.97 & 85.33\\
		\cline{2-6}
		OTE 	& STAGE-1D  & \underline{80.68} & \textbf{86.36} & 79.47	& \underline{85.97} \\
		& STAGE-2D  & \textbf{80.82} & \underline{86.02} & \underline{79.57}	& \textbf{86.51} \\
		& STAGE-3D 	& 80.39 & 85.70 & \textbf{80.03} & 85.72\\
		\midrule[1.5pt]	
		& GTS-BERT	& 66.46 & 74.63  & 67.52  & 74.20\\
		& EMC-GCN   & 67.94 & 76.33  & 67.26  & 74.15 \\
		\cline{2-6}
		AOPE 	& STAGE-1D  	& \underline{69.80} & \underline{77.47} & \underline{70.80}	& \underline{79.49} \\
		& STAGE-2D  	& \textbf{69.98} & 77.42 & \textbf{71.35}	& 79.41 \\
		& STAGE-3D 		& 69.70 & \textbf{77.87} & 70.60 & \textbf{79.98}\\
		\bottomrule
	\end{tabular}
	\caption{\label{Table:subtask_experiment}
		Test $F_1$ scores on ATE, OTE and AOPE tasks, with the best in bold and second best underlined.  
	}
\end{table}

\begin{table}[!t]\footnotesize
	\centering
	\begin{tabular}{c|c|cccc}
		\toprule
		\textbf{Setting} & 
		\textbf{Model} & \textbf{14lap} & \textbf{14res} & \textbf{15res} & \textbf{16res} \\
		\midrule[1pt]
		& GTS-BERT	& 63.33 & 75.67 & 67.51 & 72.21 \\
		& EMC-GCN 	& 64.88 & 77.88	& 68.59 & 71.64 \\
		& Span-ASTE & 67.85	& 77.25 & 69.61 & 74.11 \\
		\cline{2-6}
		\textit{Single.}	& STAGE-1D  & 70.25 & 79.67 & 70.53 & \textbf{78.11}\\
		& STAGE-2D  & \textbf{70.71} & 80.17 & \textbf{71.21} &  76.98 \\
		& STAGE-3D 	& {70.35} & \textbf{80.40} & {70.69} & {77.30} \\
		\midrule[1pt]
		& GTS-BERT	& 45.16 & 52.61 & 50.58	& 57.27 \\
		& EMC-GCN 	& 48.23 & 57.87 & 49.37	& 59.22 \\
		& Span-ASTE & {49.10} & 57.02 & {53.08}	& {62.50} \\
		\cline{2-6}
		\textit{Multi.} & STAGE-1D 	& 49.73 & \textbf{60.24} 	& 54.53 	&  63.40 \\
		& STAGE-2D  	& \textbf{50.00} 	& 59.08 	& 54.70 	&  63.55 \\
		& STAGE-3D 		& {49.50} 	& {59.20} 	& \textbf{55.25}	& \textbf{64.88}\\
		\midrule[1pt]	
		& GTS-BERT	& 48.22 & 52.79 & 54.35	& 63.29 \\
		& EMC-GCN 	& 51.57 & \textbf{59.82} & 51.98 & 64.42 \\
		& Span-ASTE & {52.24} & {59.22}	& \textbf{58.41} & {66.62} \\
		\cline{2-6}
		\textit{Multi. A.} 	& STAGE-1D  & 52.87 & 59.51 & 56.87 &  70.26 \\
					& STAGE-2D  & \textbf{53.05} & 58.86 & 57.98 &  68.98\\
					& STAGE-3D 	& {52.75} & {59.20} & {57.32} & \textbf{71.48}\\
		\midrule[1pt]
		& GTS-BERT	& 28.44 & {46.89} & {40.27} & 32.98 \\
		& EMC-GCN 	& \textbf{31.86} & 47.07	& 38.52	& 35.66 \\
		& Span-ASTE & {29.38} & 45.02 & 38.37 & \textbf{41.17} \\
		\cline{2-6}
		\textit{Multi. O.}	 & STAGE-1D & 29.30 & \textbf{57.45} & 45.47 &  33.26 \\
		& STAGE-2D & 31.18 & 53.90 & 43.55 &  38.43 \\
		& STAGE-3D  & 28.52	& {54.34} & \textbf{46.60} & {37.44} \\
		\bottomrule
	\end{tabular}
	\caption{\label{Table:additional_experiment}
		Test $F_1$ scores of ASTE under four settings, with the best in bold. \textit{Single.} denotes that triplets with single-word aspect and opinion. \textit{Multi.} denotes that triplets with multi-word aspects or multi-word opinions. \textit{Multi. A.}/\textit{Multi. O.} denotes triplets with multiple-word aspects/opinions.
	}
\end{table}
\subsection{Experiments on Subtasks}
To further investigate the effectiveness of STAGE, we conduct experiments on three subtasks of ASTE, that is, aspect term extraction (ATE), opinion term extraction (OTE) and aspect-opinion pair extraction (AOPE). Note that Our method can directly address those subtasks without additional modifications. 
Specifically, we compare our method with two state-of-the-art tagging-based methods GTS-BERT~\cite{wu-etal-2020-grid} and EMC-GCN~\cite{chen-etal-2022-enhanced} by using their official released codes and configurations, for a fair comparison. 
The results are shown in Table~\ref{Table:subtask_experiment} and the observations are that: 
(1) On ATE and OTE task, our models achieve the best results consistently on four datasets, which indicates that considering the complete semantics of each span via span tagging scheme, our method benefits the extraction of aspect and opinion terms.
(2) On AOPE task, our models also exceed the compared models by a large margin. It proves that not only our span tagging scheme can better capture aspect/opinion terms, but the greedy inference that considers the span-level constrains can improve the pairing performance as well.

\subsection{Additional Experiments}
We also compare the performance of STAGE with previous models GTS-BERT~\cite{wu-etal-2020-grid}, EMC-GCN~\cite{chen-etal-2022-enhanced} and Span-ASTE~\cite{xu-etal-2021-learning} for the following four settings : 1) Single-Word: Both aspect and opinion in a triplet are single-word terms. 2) Multi-Word: At least one of the aspect or opinion in a triplet is a multi-word term. 3) Multi-word Aspect: Aspect in a triplet is a multi-word term. 4) Multi-word Opinion: Opinion in a triplet is a multi-word term. 
The results are shown in Table~\ref{Table:additional_experiment}. The main observations are that: 
(1) Generally, compared to the three baselines, our models show superiority under all settings.
(2) Compared with Single-Word setting, Multi-Word setting poses challenges in the ASTE task, as the performances of all models drop dramatically. However, considering span-level information, Span-ASTE and our models perform better than GTS-BERT and EMC-GCN, which heavily rely on word-level interactions. 
It indicates the significance of modeling span-level information in the ASTE task.

\section{Analysis}\label{section:analysis}

\begin{table}[tb]\footnotesize
	\centering
	\begin{tabular}{c|ccc}
		\toprule
		Model & \multicolumn{3}{c}{SciERC} \\
		\cline{2-4}
		~ & \textit{Ent} & \textit{Rel} & \textit{Rel+} \\
		\hline 
		UniRE~\cite{wang-etal-2021-unire} & 68.4 & -    & 36.9 \\
		\hline 
		PURE-S~\cite{zhong-chen-2021-frustratingly} & 66.6 & 48.2 & 35.6 \\
		PURE-C~\cite{zhong-chen-2021-frustratingly} $^\dag$ & 68.9 & \textbf{50.1} & 36.8 \\
		\hline 
		PFN~\cite{yan-etal-2021-partition} & 66.8 & - & 38.4 \\
		\hline
		STAGE & \textbf{69.1} & 48.8 & \textbf{38.9} \\
		\hline
		
	\end{tabular}
	\caption{
		\label{Table:scierc_result}
		Test $F_1$ scores on SciERC dataset with the best in bold. $^\dag$ means using cross-sentence context. \textit{Ent} denotes the entity boundaries evaluation. \textit{Rel} denotes the entity boundaries and relation being correct and \textit{Rel+} denotes the eneity boundaries, entity types and the realtion being correct.
	}
\end{table}
\subsection{Scheme Generalization}
In this part, we extend our STAGE in entity and relation extraction task.
Similar to the ASTE task, the span in the given sentence may have two roles: an entity term or a relation snippet that includes a pair of entities with a specific relation, forming its boundaries. So, we can define two role dimensions (entity and relation snippet) for entity and relation extraction task.
But unlike the ASTE task, the ``relation'' here has ``direction'' (\ie from head entity to tail entity). 
Thus, we adopt the span tagging scheme with the following tag set:
$\{N, E_1, ..., E_{|E|}\}$ - $\{N, R^{(H-T)}_1, R^{(T-H)}_1, ..., R^{(T-H)}_{|R|}\}$,
where $E = \{E_1, E_2, ..., E_{|E|}\}$ denotes entity categories, 
$R = \{R_1,R_2,...,R_{|R|}\}$ denotes relation types, $N$ means nothing in that role dimension,
${(H-T)}$ and $(T-H)$ discriminate the relation directions as the former indicates head entity appears before the tail entity and the latter shows the opposite, and  $|\cdot|$ represents the set size.

We conduct experiments on SciERC dataset~\cite{luan-etal-2018-multi}, a widely-used dataset including 500 scientific articles with annotations on the scientific terms and their relations. 
We follow the experiment settings of previous work~\cite{zhong-chen-2021-frustratingly} and utilize micro F1 score~($F_1$) as the evaluation metric. 
The experimental results in Table~\ref{Table:scierc_result} show our model also achieves good performance, which verifies the good generalization ability of STAGE.

\section{Conclusion}
In this paper, we propose a novel approach,  \underline{\textbf{S}}pan \underline{\textbf{TA}}gging and  \underline{\textbf{G}}reedy inf\underline{\textbf{E}}rence (STAGE), to solve the ASTE task in an end-to-end manner.
To the best of our knowledge, it is the first effort to explore a span-level tagging method and formulate the ASTE task as a multi-class \textbf{span} classification problem.
Naturally performing on \textit{span-level}, STAGE can overcome the limitations of previous tagging methods and effectively generate more accurate sentiment triplets by exploring span-level information and constraints. 
Specifically, STAGE consists of two components, namely \textit{span tagging} scheme and \textit{greedy inference} strategy. 
The former considers the diversity of span roles and tags the span based on three pre-defined role dimensions. 
The latter considers the mutual span-level constraints and retrieves the aspect/opinion term with the maximum length from the candidate sentiment snippet.
Based on STAGE, we propose a simple but effective model, which outperforms the state-of-the-arts by a large margin on four widely-used datasets. 
Moreover, it can be easily generalized to other pair/triplet extraction tasks, indicating the superiority of the scheme STAGE.

\section{Acknowledgments}
This work was supported in part by the National Natural Science Foundation of China under Grant No. 62276110, Grant No.61772076, in part by CCF-AFSG Research Fund under Grant No.RF20210005, and in part by the fund of Joint Laboratory of HUST and Pingan Property \& Casualty Research (HPL). The authors would also like to thank the anonymous reviewers for their comments on improving the quality of this paper. 

\bibliography{aaai23} 
\clearpage
\appendix
\begin{table*}[htb]\footnotesize
	\centering
	\begin{tabular}{c|cccc}
		\toprule
		~ & \textbf{\textit{Sentence:}} & \multicolumn{3}{l}{\normalsize{The menu is interesting and quite reasonably priced .}} \\
		\cline{2-5}
		~ & \textbf{Golden} & (menu, interesting, POS) 
		& (menu, reasonably priced, POS)
		& (priced, reasonably, POS) \\
		~ & \textbf{GTS-BERT} & \textsurd & (menu, \textbf{\textit{reasonably}}, POS) &  \textsurd \\
		~ & \textbf{EMC-GCN} & \textsurd & (menu, \textbf{\textit{reasonably}}, POS) &  \textsurd \\
		(a) & \textbf{Span-ASTE} & \textsurd & \textbf{\texttimes} &  \textsurd \\
		\cline{2-5}
		~ & \textbf{STAGE-1D} & \textsurd & \textsurd &  \textbf{\texttimes} \\
		~ & \textbf{STAGE-2D} & \textsurd & \textsurd & \textsurd  \\
		~ & \textbf{STAGE-3D} & \textsurd & \textsurd & \textsurd  \\
		\midrule
		~ & \textbf{\textit{Sentence:}} & \multicolumn{3}{l}{\normalsize{BEST spicy tuna roll , great asian salad .}} \\
		\cline{2-5}
		~ & \textbf{Golden} & (spicy tuna roll, BEST, POS) 
		& (asian salad, great, POS)
		& ~ \\
		~ & \textbf{GTS-BERT} & (\textbf{\textit{tuna roll}}, BEST, POS) & \textsurd & ~ \\
		~ & \textbf{EMC-GCN}  & (\textbf{\textit{tuna roll}}, BEST, POS) & \textsurd &  ~ \\
		(b) & \textbf{Span-ASTE} & \textsurd & \textsurd &  (\textbf{\textit{tuna roll}}, BEST, POS) \\
		\cline{2-5}
		~ & \textbf{STAGE-1D} & \textsurd & \textsurd & ~ \\
		~ & \textbf{STAGE-2D} & \textsurd & \textsurd & ~  \\
		~ & \textbf{STAGE-3D} & \textsurd & \textsurd & ~  \\
		\midrule
		~ & \textbf{\textit{Sentence:}} & \multicolumn{3}{l}{\normalsize{Service is excellent , no wait , and you get a lot for the price .}} \\
		\cline{2-5}
		~ & \textbf{Golden} & (Service, excellent, POS) & (wait, no, POS)& ~ \\
		~ & \textbf{GTS-BERT} & \textsurd & (wait, no, \textbf{\textit{NEG}}) & ~ \\
		~ & \textbf{EMC-GCN}  & \textsurd & (wait, no, \textbf{\textit{NEG}}) &  ~ \\
		(c) & \textbf{Span-ASTE} & \textsurd & \textbf{\texttimes} & ~\\
		\cline{2-5}
		~ & \textbf{STAGE-1D} & \textsurd & \textsurd & ~ \\
		~ & \textbf{STAGE-2D} & \textsurd & \textsurd & ~  \\
		~ & \textbf{STAGE-3D} & \textsurd & \textsurd & ~  \\
		
		\bottomrule
	\end{tabular}
	\vspace{-4pt}
	\caption{\label{Table:qualitative_analysis} Example cases with golden labels and predictions from three baselines and our models. \textsurd \ means a correct prediction,  \textbf{\texttimes} represents failing to generate. Incorrect elements of predicted triplets are shown in \textbf{\textit{bold}}.}
	\vspace{-4pt}
\end{table*}

\begin{table*}[!t]\footnotesize
	\centering
	\begin{tabular}{cc|c|ccc|ccc|ccc}
		\toprule
		\multicolumn{2}{c|}{\textbf{Dataset}} & Total 
		& \multicolumn{3}{c|}{\textbf{Types of Limitations}} 
		& \multicolumn{3}{c|}{\textbf{Types of Multiple Roles}} 
		& \multicolumn{3}{c}{\textbf{\# Fail cases}} \\
		~ & ~ & triplets & \textit{Nested.} & \textit{Conflict.} & \textit{Greedy.} 
		& \textit{A-O-N}   & \textit{A/O-P}     & \textit{A-O-P}
		& \textit{3D}      & \textit{2D}        & \textit{1D}\\
		\midrule
		~ 	  & Train & 1460 & 1  & 0  & 0  & 2 & 5  & 0 & 1  & 3 & 8  \\
		14lap & Dev   & 345  & 0  & 1  & 0  & 0 & 0  & 0 & 1  & 1 & 1  \\
		~ 	  & Test  & 541  & 0  & 0  & 0  & 0 & 1  & 0 & 0  & 0 & 1  \\
		\hline
		~     & Train & 2337 & 3  & 0  & 1  & 0 & 20 & 1 & 4  & 5 & 25 \\
		14res & Dev   & 577  & 0  & 0  & 0  & 0 & 8  & 0 & 0  & 0 & 8  \\
		~     & Test  & 994  & 0  & 0  & 1  & 1 & 10 & 0 & 1  & 2 & 12 \\
		\hline
		~     & Train & 1013 & 0  & 0  & 0  & 0 & 1  & 0 & 0  & 0 & 1  \\
		15res & Dev   & 249  & 0  & 0  & 0  & 0 & 0  & 0 & 0  & 0 & 0  \\
		~     & Test  & 485  & 0  & 0  & 1  & 0 & 0  & 0 & 1  & 1 & 1  \\
		\hline
		~     & Train & 1394 & 0  & 0  & 1  & 0 & 1  & 0 & 1  & 1 & 2  \\
		16res & Dev   & 339  & 0  & 0  & 0  & 0 & 0  & 0 & 0  & 0 & 0  \\
		~     & Test  & 514  & 0  & 0  & 0  & 1 & 1  & 0 & 0  & 1 & 2  \\
		\midrule 
		\multicolumn{2}{c|}{\textit{Sum.}} & 10248 & 4 & 1 & 4 & 4 & 47 & 1 & 9 & 14 & 61 \\
		\bottomrule
	\end{tabular}
	\vspace{-2pt}
	\caption{\label{Table:analysis_tagging_scheme_limitations} More \textbf{triplets} statistics. The \textit{Nested.} denotes that the aspect(opinion) term is completely nested with its opinion(aspect) term. The \textit{Conflict.} means multiple triplets correspond to the same sentiment snippet. \textit{Greedy.} represents a longer irrelevant term is wrongly selected due to the greedy inference method. The \textit{A-O-N} means a span being aspect term and opinion term simultaneously. The \textit{A/O-P} means a span being aspect(opinion) and sentiment snippet simultaneously. The \textit{A-O-P} denotes a span being aspect term , opinion term, and sentiment snippet simultaneously. \textit{3D}, \textit{2D}, \textit{1D} represents our 3D-version, 2D-version, 1D-version span tagging schemes, respectively. \textit{Sum.} means the sum of triplets in the same type.}
	\vspace{-2pt}
\end{table*}

\begin{table*}[!t]\footnotesize
	\centering
	\begin{tabular}{c|ccc|ccc|ccc|ccc}
		\toprule
		\textbf{Dataset} & \multicolumn{3}{c|}{\textbf{14lap}} & \multicolumn{3}{c|}{\textbf{14res}} & \multicolumn{3}{c|}{\textbf{15res}} & \multicolumn{3}{c}{\textbf{16res}} \\
		
		~  & Train & Dev & Test &  TraHin & Dev & Test  & Train & Dev & Test &  Train & Dev & Test \\
		\midrule
		\textbf{Single.}	& 824 & 190 & 289 & 1585 & 388 & 657
		& 678 & 165 & 297 & 918 & 216 & 344 \\
		\textbf{Multi.} 	& 636 & 155 & 252 & 752 & 189 & 337
		& 335 & 84  & 188 & 476 & 123 & 344  \\
		\textbf{Multi. A.}  & 497 & 120 & 220 & 560 & 147 & 269 
		& 258 & 63  & 137 & 358 & 94  & 129\\
		\textbf{Multi. O.}  & 228 & 62  & 69  & 259 & 53  & 91
		& 107 & 33  & 64  & 164 & 38  & 54\\
		
		\bottomrule
	\end{tabular}
	\caption{\label{Table:detail_dataset_statistics} More detailed Statistics of datasets. ``Single.'' is the number of triplets that both the aspect term and opinion term contain only one word. ``Multi.'' denotes the number of the triplets that at least one of the aspect and opinion term contain multiple words. ``Multi. A.'' and ``Multi. O.'' are the number of triplets containing multiple-word aspects and opinions, respectively.}
\end{table*}

\begin{table*}[!t]\footnotesize
	\centering
	\begin{tabular}{c|c|ccc|ccc|ccc|ccc}
		\toprule
		\textbf{Task} & 
		\textbf{Model} & \multicolumn{3}{c|}{\textbf{14lap}} & \multicolumn{3}{c|}{\textbf{14res}} & \multicolumn{3}{c|}{\textbf{15res}} & \multicolumn{3}{c}{\textbf{16res}} \\
		~ & ~  & $P$ & $R$ & $F_1$ & $P$ & $R$ & $F_1$ &$P$ & $R$ & $F_1$ & $P$ & $R$ & $F_1$ \\
		\midrule[1pt]
		& GTS-BERT	& 77.00 & 84.19 & 80.44
		& 78.88 & 87.05 & 82.76 
		& 75.13 & 82.78 & 78.27
		& 75.20 & 88.12 & 81.13 \\
		& EMC-GCN 	& 79.31 & 84.19 & 81.67 
		& 80.90 & 88.83 & 84.68
		& 73.53 & 82.22 & 77.62
		& 73.25 & 89.53 & 80.57 \\
		\cline{2-14}
		ATE & STAGE-1D  & 80.91 & 85.14 & \textbf{82.96}
		& 82.62 & 88.07 & \textbf{85.24}
		& 79.65 & 82.27 & 80.91
		& 79.89 & 87.70 & 83.60\\
		& STAGE-2D  & 80.45 & 85.10 & \underline{82.64}
		& 82.01 & 88.73 & \underline{85.23} 
		& 80.22 & 82.64 & \textbf{81.39}
		& 80.90 & 87.26 & \textbf{83.95}\\
		& STAGE-3D 	& 80.29 & 85.10 & {82.62}
		& 81.43 & 88.89 & {84.99}
		& 81.63 & 80.83 & \underline{81.21}
		& 79.02 & 89.38 & \underline{83.86} \\
		\midrule[1.5pt]
		& GTS-BERT	& 76.10 & 79.62 & 77.82
		& 80.91 & 89.22 & 84.85 
		& 72.97 & 82.74 & 77.53
		& 80.36 & 88.79 & 84.36\\
		& EMC-GCN 	& 78.08 & 79.70 & 78.85 
		& 82.12 & 89.45 & 85.62 
		& 75.59 & 82.70 & 78.97 
		& 80.53 & 90.74 & 85.33 \\
		\cline{2-14}
		OTE & STAGE-1D  &78.10 & 83.54 & \underline{80.68} & 83.43 & 89.53 & \textbf{86.36} & 76.08 & 83.21 & 79.47 & 82.36 & 89.94 & \underline{85.97} \\
		& STAGE-2D  & 78.85 & 82.95 & \textbf{80.82} & 82.93 & 89.39 & \underline{86.02} & 76.58 & 82.86 & \underline{79.57} & 83.24 & 90.06 & \textbf{86.51} \\
		& STAGE-3D 	& 78.95 & 81.94 & 80.39 & 82.31 & 89.41 & 85.70 & 76.99 & 83.34 & \textbf{80.03} & 81.45 & 90.48 & 85.72 \\
		\midrule[1.5pt]	
		& GTS-BERT & 70.45 & 62.92 & 66.46 & 74.30 & 74.98 & 74.63 & 69.07 & 66.10 & 67.52& 72.79 & 75.71 & 74.20 \\
		& EMC-GCN 	& 69.95 & 66.06 & 67.94 & 75.04 & 77.68 & 76.33 & 66.06 & 68.54 & 67.26& 69.49 & 79.57 & 74.15 \\
		\cline{2-14}
		AOPE & STAGE-1D  	& 81.32 & 61.18 & \underline{69.80} & 83.74 & 72.09 & \underline{77.47} & 79.25 & 64.04 & \underline{70.80} & 84.83 & 74.79 & \underline{79.49} \\
		& STAGE-2D  	& 79.94 & 62.28 & \textbf{69.98} & 82.57 & 72.88 & 77.42 & 79.46 & 64.74 & \textbf{71.35} & 84.78 & 74.71 & 79.41\\
		& STAGE-3D 		& 81.62 & 60.88 & 69.70 & 82.94 & 73.46 & \textbf{77.87} & 80.23 & 63.09 & 70.60 & 83.73 & 76.58 & \textbf{79.98}\\
		\bottomrule
	\end{tabular}
	\caption{\label{Table:subtask_experiment_appendix} Experimental results on the ATE, OTE and AOPE tasks with best $F_1$ in \textbf{bold} and the second best are \underline{underlined}.}
\end{table*}

\begin{table*}[!t]\footnotesize
	\centering
	\begin{tabular}{c|c|ccc|ccc|ccc|ccc}
		\toprule
		\textbf{Task} & 
		\textbf{Model} & \multicolumn{3}{c|}{\textbf{14lap}} & \multicolumn{3}{c|}{\textbf{14res}} & \multicolumn{3}{c|}{\textbf{15res}} & \multicolumn{3}{c}{\textbf{16res}} \\
		
		~ & ~  & $P$ & $R$ & $F_1$ & $P$ & $R$ & $F_1$ &$P$ & $R$ & $F_1$ & $P$ & $R$ & $F_1$  \\
		\midrule[1pt]
		& GTS-BERT	& 64.39 & 62.34 & 63.33 & 73.12 & 78.41 & 75.67 & 68.54 & 66.55 & 67.51 & 70.05 & 74.55 & 72.21  \\
		& EMC-GCN 	& 61.80 & 68.34 & 64.88 & 74.74 & 81.31 & 77.88 & 65.07 & 72.57 & 68.59 & 66.23 & 78.03 & 71.64 \\
		& Span-ASTE & 68.14 & 67.61 & 67.85 & 77.05 & 77.66 & 77.25 & 71.80 & 67.61 & 69.61 & 72.89 & 75.41 & 74.11 \\
		\cline{2-14}
		Single.	& STAGE-1D  & 76.33 & 65.09 & 70.25 & 82.69 & 76.93 & 79.69 & 78.37 & 64.18 & 70.53 & 81.49 & 75.00 & \textbf{78.11} \\
		& STAGE-2D  & 75.48 & 66.53 & \textbf{70.71} & 81.75 & 78.66 & \underline{80.17} & 78.08 & 65.45 & \textbf{71.21} & 80.14 & 74.07 & 76.98 \\
		& STAGE-3D 	& 75.82 & 65.70 & \underline{70.35} & 82.19 & 78.72 & \textbf{80.40} & 79.34 & 63.77 & \underline{70.69} & 80.42 & 74.42 & \underline{77.30}  \\
		\midrule[1pt]
		& GTS-BERT	& 50.61 & 40.80 & 45.16 & 56.16 & 49.54 & 52.61 & 52.41 & 48.99 & 50.58 & 57.69 & 57.02 & 57.27  \\
		& EMC-GCN 	& 55.97 & 42.39 & 48.23 & 60.10 & 55.84 & 57.87 & 51.70 & 47.30 & 49.37 & 57.32 & 61.55 & 59.22  \\
		& Span-ASTE & 55.11 & 44.29 & 49.10 & 66.70 & 49.97 & 57.02 & 53.01 & 53.40 & 53.08 & 62.00 & 63.06 & 62.50  \\
		\cline{2-14}
		Multi. & STAGE-1D 	& 64.03 & 40.71 & \underline{49.73} & 71.67 & 51.99 & \textbf{60.24} & 61.87 & 48.83 & 54.53 & 71.19 & 57.18 & 63.40 \\
		& STAGE-2D  	& 62.97 & 41.59 & \textbf{50.00} & 70.31 & 51.04 & 59.08 & 62.56 & 48.62 & \underline{54.70} & 71.92 & 57.06 & \underline{63.55} \\
		& STAGE-3D 		& 65.70 & 39.76 & 49.50 & 69.67 & 51.75 & \underline{59.20} & 64.18 & 48.62 & \textbf{55.25} & 68.84 & 61.41 & \textbf{64.88} \\
		\midrule[1pt]	
		& GTS-BERT	& 54.47 & 43.27 & 48.22 & 55.93 & 50.04 & 52.79 & 55.58 & 53.28 & 54.34 & 61.86 & 64.84 & 63.29  \\
		& EMC-GCN 	& 59.32 & 45.64 & 51.57 & 61.23 & 58.54 & \textbf{59.82} & 52.10 & 51.97 & 51.98 & 60.19 & 69.69 & 64.42 \\
		& Span-ASTE & 57.88 & 47.64 & 52.24 & 68.07 & 52.57 & 59.22 & 59.33 & 57.66 & \textbf{58.41} & 63.98 & 69.61 & 66.62  \\
		\cline{2-14}
		Multi. A. 	& STAGE-1D  & 66.74 & 43.82 & \underline{52.87} & 71.43 & 51.08 & \underline{59.51} & 66.60 & 49.78 & 56.87 & 75.93 & 65.43 & \underline{70.26} \\
		& STAGE-2D  & 65.95 & 44.55 & \textbf{53.05} & 70.35 & 50.71 & 58.86 & 67.80 & 50.66 & \underline{57.98} & 76.02 & 63.26 & 68.98 \\
		& STAGE-3D 	& 68.16 & 43.09 & 52.75 & 70.12 & 51.60 & 59.20 & 68.31 & 49.49 & 57.32 & 74.32 & 68.99 & \textbf{71.48} \\
		\midrule[1pt]
		& GTS-BERT	& 31.06 & 26.29 & 28.44 & 52.31 & 42.64 & 46.89 & 43.16 & 37.85 & 40.27 & 35.22 & 31.54 & 32.98 \\
		& EMC-GCN 	& 38.15 & 27.35 & \textbf{31.86} & 52.54 & 42.73 & 47.07 & 47.65 & 32.62 & 38.52 & 37.52 & 34.23 & 35.66  \\
		& Span-ASTE & 35.28 & 25.22 & 29.38 & 61.04 & 36.48 & 45.02 & 38.24 & 39.06 & 38.37 & 43.45 & 39.26 & \textbf{41.17} \\
		\cline{2-14}
		Multi. O.	 & STAGE-1D & 39.67 & 23.48 & 29.30 & 69.52 & 49.01 & \textbf{57.45} & 50.26 & 41.56 & \underline{45.47} & 45.44 & 26.30 & 33.26 \\
		& STAGE-2D  & 41.16 & 25.22 & \underline{31.18} & 65.77 & 45.93 & 53.90 & 49.36 & 39.06 & 43.55 & 49.97 & 31.48 & \underline{38.43}\\
		& STAGE-3D 	& 44.84 & 21.16 & 28.52 & 65.24 & 46.59 & \underline{54.34} & 53.39 & 41.56 & \textbf{46.60} & 44.09 & 32.59 & 37.44  \\
		\bottomrule
	\end{tabular}
	\vspace{-2pt}
	\caption{\label{Table:additional_experiment_appendix}
		Additional Experimental results on four different settings with best $F_1$ in \textbf{bold} and the second best are \underline{underlined}. 
	}
	\vspace{-2pt}
\end{table*}
\section{Qualitative Analysis}
In this part, we present sample sentences from the real-word ASTE test dataset with the golden labels as well as predictions from three baselines and our models, which are shown in Figure~\ref{Table:qualitative_analysis}. The observations are that:

(1) \textbf{\textit{Multiple roles of a word}} and \textbf{\textit{Multiple relations of a word pair}} challenge the previous tagging-based models. In sentence(a), ``priced'' has two roles, \ie being (part of) an aspect term and an opinion term and the word pair ``priced-reasonably'' has two relations, \ie belonging to the same opinion and forming a valid sentiment pair. Both GTS-BERT and EMC-GCN fail to output correct triplets.

(2) \textbf{\textit{Multi-word term}} also causes baseline models to perform poorly. In sentence (b), both GTS-BERT and EMC-GCN extract the wrong term ``tuna roll'' instead of ``spicy tuna roll''. 
It may result from the significant semantic difference between the aspect ``spicy tuna roll'' and ``spicy'', and treating a multi-word term as independent word sets, previous \textit{word-level} tagging scheme fail to extract the multi-word aspect/opinion term preciously. 
On the contrary, Span-ASTE and our models, which consider \textit{span-level} information, extract the multi-word term correctly. 

(3) \textbf{\textit{Greedy inference}} benefits the triplet extraction. 
In sentence (b), Span-ASTE, which considers learning span-span interactions, wrongly extracts the triplet ``(tuna roll, BEST, POS)'', even though it already predicts the correct one ``(spicy tuna roll, BEST, POS)''. 
That is because Span-ASTE adopts the commonly used decoding process that looks into every candidate aspect-opinion pair combination to see if valid, without considering the influence between candidate triplets. 
On the contrary, even if both two triplets are the candidates, we would only retrieve the aspect/opinion term with the maximum length ( \ie ``spicy tuna roll'') under the guidance of the \textit{Greedy inference} strategy and generate the exact one correct triplet ``(spicy tuna roll, BEST, POS)''.

(4) \textbf{\textit{Sentiment snippet}} can learn the correct semantic combination of the aspect and opinion term. In sentence (c), both GTS-BERT and EMC-GCN wrongly predict the sentiment of the pair (wait, no). The possible reason is that the ``wait'' and ``no'' both convey negativeness, and by only considering their independent representations, the above two baselines fall short of learning their semantic combination correctly. However, our STAGE views them as a whole span with a specific role (\ie sentiment snippet) and performs better.

\begin{figure}[htb]
	\centering
	\includegraphics[width=0.48\textwidth]{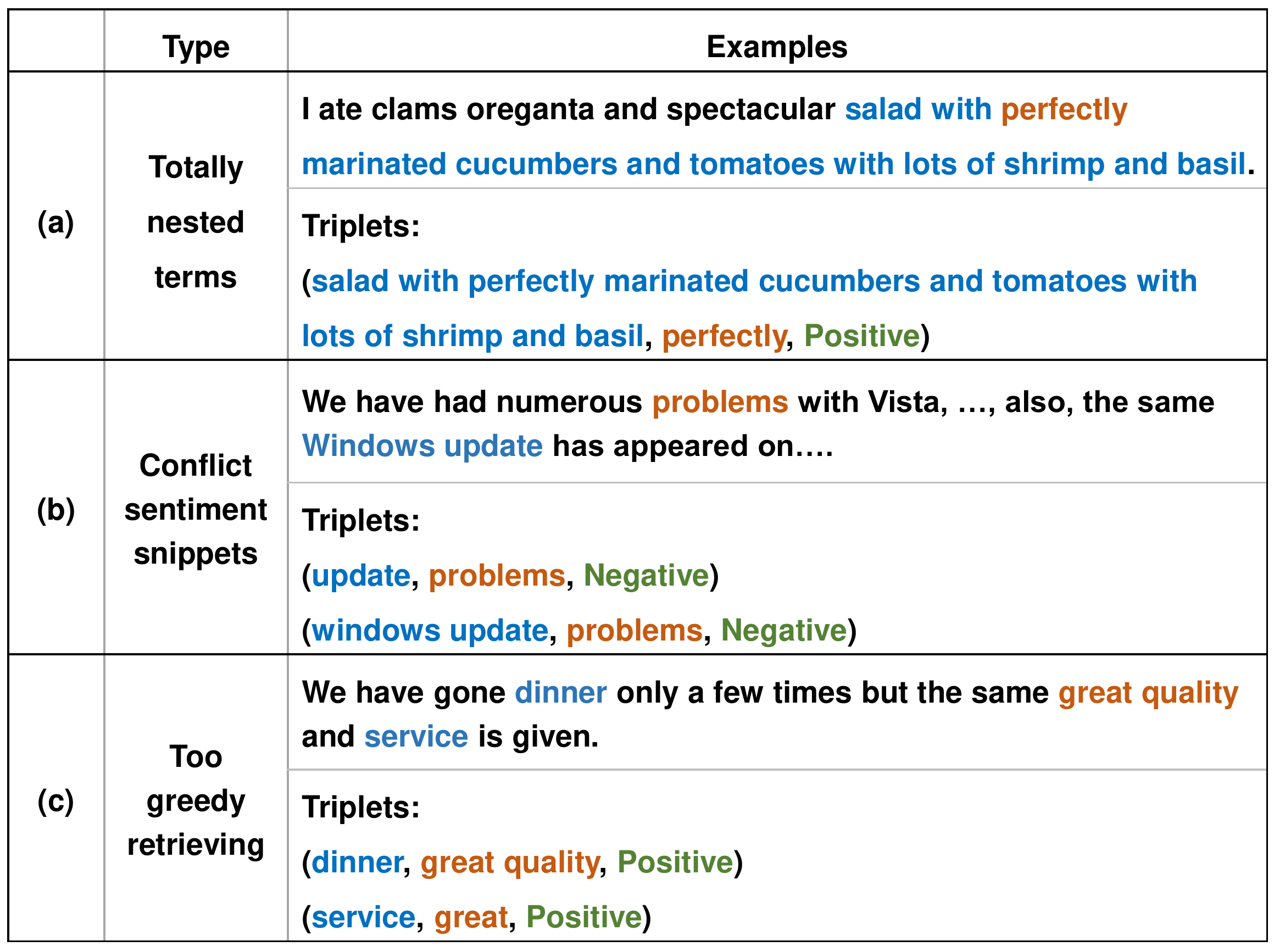}
	\caption{\label{Fig:fail_case}Failure cases.}
	\vspace{-8pt}
\end{figure}

\section{Limitations}
To fully understand our scheme STAGE, we also analyze its limitations. 
Although STAGE achieves state-of-the-art performance in extensive experimental settings and more complex situations, it still fails in some cases\footnote{We want to point out that all previous tagging schemes cannot handle those situations as well.}. 

(1) \textbf{\textit{Totally nested terms}}: The \textit{span tagging} scheme fails to find proper tags when an aspect~(opinion) term is totally nested (with no shared boundaries) in the corresponding opinion~(aspect) term, which is exemplified by Figure~\ref{Fig:fail_case}~(a).


(2) \textbf{\textit{Conflict sentiment snippets}}: STAGE also cannot handle the situation that a sentiment snippet corresponds to multiple aspect-opinion pairs, as Figure~\ref{Fig:fail_case}~(b) shows. 

(3) \textbf{\textit{Too greedy retrieving}}: The \textit{greedy inference} fails to output correct triplets when other longer aspect/opinion term appears in the sentiment snippet and shares the boundary just as its counterpart does, as exemplified by Figure~\ref{Fig:fail_case}~(c). 
Even though correct tag results are generated, the greedy inference for the ``great quality and service'' can only extract ``(service, great quality, Positive)'' because opinion term ``great quality'' is longer than the golden opinion term ``great''.

Considering these cases are rare (as Table~\ref{Table:analysis_tagging_scheme_limitations} shows), we leave the perfection of STAGE as the future work. 

\section{Experimental Details}

\subsection{Additional Data Statistics}
Table~\ref{Table:detail_dataset_statistics} contains more details about the four datasets. It shows that multi-word triplets are common as they make up a third of the samples. Besides, multi-word opinions are much less than multi-word aspects, which can explain the inferior experiment results in ``Multi-word Opinion'' setting, compared with ``Multi-word Aspect'' setting.
\subsection{Additional Experimental Settings}
We conduct experiments on a Nvidia GeForce 3090 GPU, with CUDA 11.6 and PyTorch 1.10.1. The average run time for 14lap, 14res, 15res and 16res is 2.7, 3.6 ,1.8 and 2.4 sec/epoch. Total number of parameters is about 110M.

\subsection{Detailed Experimental Results}
Table~\ref{Table:subtask_experiment_appendix} shows detailed results in ATE, OTE and AOPE subtasks. 
Table~\ref{Table:additional_experiment_appendix} shows more scores on four different settings.

\end{document}